\documentclass[sigconf, screen]{acmart}

\usepackage{xcolor}
\usepackage{booktabs}
\usepackage{graphicx}
\usepackage{multirow}
\usepackage{hyperref}
\usepackage{float}
\usepackage{algorithm}
\usepackage{algpseudocode}
\AtBeginDocument{%
  }

\setcopyright{acmlicensed}
\copyrightyear{2026}
\acmYear{2026}
\acmDOI{XXXXXXX.XXXXXXX}
\acmConference[MM '26]{The 34th ACM International Conference on Multimedia}{November 10--14, 2026}{Rio de Janeiro, Brazil}
\acmISBN{978-1-4503-XXXX-X/2018/06}

\acmSubmissionID{1693}

\begin{document}
\renewcommand\footnotetextcopyrightpermission[1]{}
\settopmatter{printacmref=false} 
\title{Challenging Vision-Language Models with Physically Deployable Multimodal Semantic Lighting Attacks}

\author{Yingying Zhao, Chengyin Hu, Qike Zhang, Xin Li, Xin Wang, Yiwei Wei,
Jiujiang Guo, Jiahuan Long, Tingsong Jiang, Wen Yao}

\renewcommand{\shortauthors}{Trovato et al.}

\begin{abstract}
Vision-Language Models (VLMs) have shown remarkable performance, yet their security remains insufficiently understood. Existing adversarial studies focus almost exclusively on the digital setting, leaving physical-world threats largely unexplored. As VLMs are increasingly deployed in real environments, this gap becomes critical, since adversarial perturbations must be physically realizable. Despite this practical relevance, physical attacks against VLMs have not been systematically studied. Such attacks may induce recognition failures and further disrupt multimodal reasoning, leading to severe semantic misinterpretation in downstream tasks. Therefore, investigating physical attacks on VLMs is essential for assessing their real-world security risks. To address this gap, we propose Multimodal Semantic Lighting Attacks (MSLA), the first physically deployable adversarial attack framework against VLMs. MSLA uses controllable adversarial lighting to disrupt multimodal semantic understanding in real scenes, attacking semantic alignment rather than only task-specific outputs. Consequently, it degrades zero-shot classification performance of mainstream CLIP variants while inducing severe semantic hallucinations in advanced VLMs such as LLaVA and BLIP across image captioning and visual question answering (VQA). Extensive experiments in both digital and physical domains demonstrate that MSLA is effective, transferable, and practically realizable. Our findings provide the first evidence that VLMs are highly vulnerable to physically deployable semantic attacks, exposing a previously overlooked robustness gap and underscoring the urgent need for physical-world robustness evaluation of VLMs.
\end{abstract}



\maketitle

\section{Introduction}
In recent years, classical Vision-Language Models (VLMs)~\cite{ref1,ref2} have achieved cross-modal embedding alignment through large-scale image-text contrastive learning~\cite{ref23}. Meanwhile, further advances in vision-language pretraining have enabled these models to extend to generative and comprehension tasks~\cite{ref3}, such as image captioning~\cite{ref4,ref21} and visual question answering (VQA)~\cite{ref5,ref6}. Current VLMs~\cite{ref5,ref6,ref7} have achieved remarkable progress in bridging computer vision and natural language processing. Owing to their powerful cross-modal understanding, VLMs are being rapidly deployed in safety-critical real-world applications, including autonomous driving~\cite{ref34,ref51} and security surveillance~\cite{ref8,ref25}.

However, despite their strong performance across diverse tasks, VLMs remain vulnerable to adversarial examples~\cite{ref9,ref15,ref16}. In real-world scenarios, they inevitably encounter complex environmental perturbations~\cite{ref12,ref14,ref41}. Among these, illumination variation is one of the most common and unavoidable factors. Notably, most existing adversarial attacks against VLMs are limited to the digital domain. In practical deployment, however, physical adversarial attacks pose a more realistic and severe threat than digital attacks, which typically rely on imperceptible pixel-level noise~\cite{ref11}. Although several studies have investigated illumination-based physical attacks~\cite{ref17,ref18,ref19}, most of them focus on conventional deep neural networks rather than VLMs, leaving the vulnerability of VLMs to physical light perturbations largely unexplored. 

To bridge this research gap, we propose the first physical attack against VLMs, termed Multimodal Semantic Lighting Attacks (MSLA), to systematically investigate the vulnerability of VLMs to light perturbations. We model the triangular light using a 9-dimensional parameter space, including center coordinates, radius, color, and polar angles, resulting in a complex non-convex optimization problem. In this setting, traditional optimization methods are easily trapped in local optima, while genetic algorithms~\cite{ref20} can better explore the search space through crossover and mutation. Experimental results show that MSLA significantly degrades VLM performance under optimized light parameters and remains effective in real-world physical deployments, revealing the vulnerability of VLMs to realistic light-based perturbations. The main contributions of this work are summarized as follows:
\begin{itemize}
    \item To the best of our knowledge, we propose the first physical adversarial attack framework against VLMs, termed MSLA, and systematically investigate the feasibility and potential threat of exploiting localized triangular illumination to attack VLMs.
    
    \item We conduct extensive experiments in both digital and physical settings. The results demonstrate that MSLA achieves strong attack effectiveness across both domains, exposing the vulnerability of VLMs to real-world physical light perturbations. 
    
    \item Through in-depth analyses on zero-shot classification, image captioning, and VQA, we reveal how localized light perturbations interfere with visual attention and feature encoding, leading to semantic shifts and cross-modal reasoning failures in VLMs.
\end{itemize}

\section{Related Work}
\subsection{Robustness of Vision-Language Models}
Driven by large-scale multimodal pre-training, foundation models~\cite{ref1,ref2,ref35,ref36} seamlessly integrate visual and linguistic features. According to the manner of multimodal interaction, existing VLMs can generally be categorized into fused models and aligned models~\cite{ref49}. Fused models such as BLIP~\cite{ref50} employ a joint encoder to perform unified modeling of image and text information, thereby learning shared multimodal representations. In contrast, aligned models (e.g., CLIP~\cite{ref1}) use separate image and text encoders to extract features, and then achieve cross-modal alignment via contrastive learning or matching mechanisms. Based on these pre-training paradigms, recent VLMs~\cite{ref5,ref6,ref7} have demonstrated strong capabilities in visual understanding and natural language generation. However, as these models are increasingly deployed in real-world applications, research on the robustness of VLMs has also attracted increasing attention. Existing studies~\cite{ref39,ref40} have shown that VLMs still exhibit limitations in modeling fine-grained visual attributes, such as spatial relations and numerical information. Furthermore, since most VLMs remain highly dependent on the features extracted by visual encoders at the architectural level, their outputs are relatively sensitive to perturbations in input images, especially when sufficient textual context is unavailable as a constraint~\cite{ref11,ref37,ref38}.

Adversarial attacks against VLMs have attracted increasing attention~\cite{ref52,ref53}. Although prior studies have shown that VLMs are vulnerable to $L_p$-norm perturbations~\cite{ref11,ref42}, their robustness to light interference in real-world scenarios remains unexplored. To fill this gap, MSLA systematically evaluates the impact of light perturbations on VLMs in both digital and physical settings.

\subsection{Physical Adversarial Light Attacks}

To overcome the limitations of digital-domain attacks in real-world scenarios, researchers have increasingly explored physical adversarial attacks. Among them, light-based attacks are particularly stealthy and practical, as lighting is ubiquitous, difficult to control, and often appears as natural visual effects~\cite{ref13}. Specifically, several studies focus on active light attacks, where OPAD~\cite{ref18} and SPAA~\cite{ref46} use projectors to cast adversarial light onto target objects or scenes, and AdvLB~\cite{ref41} employs laser beams to generate localized high-intensity perturbations. Other works exploit ambient lighting conditions, such as manipulating shadows~\cite{ref47} or reflected light~\cite{ref19}, to create adversarial patterns. Moreover, prior research~\cite{ref48} suggests that even common natural light spots can threaten model performance.

However, most of these attacks focus on specific applications such as traffic sign recognition, and their targets are primarily CNN-based visual models rather than modern VLMs. Notably, current studies on the illumination robustness of VLMs~\cite{ref15,ref16} are generally limited to the digital domain and have not further evaluated these models in real-world physical deployment settings. In contrast, the proposed MSLA framework adopts a lightweight parameterized modeling strategy and leverages localized light, jointly constrained by geometric features and color attributes, to assess the illumination robustness of VLMs in the real physical world.

\begin{figure*}[t]
  \centering
  \includegraphics[width=\textwidth]{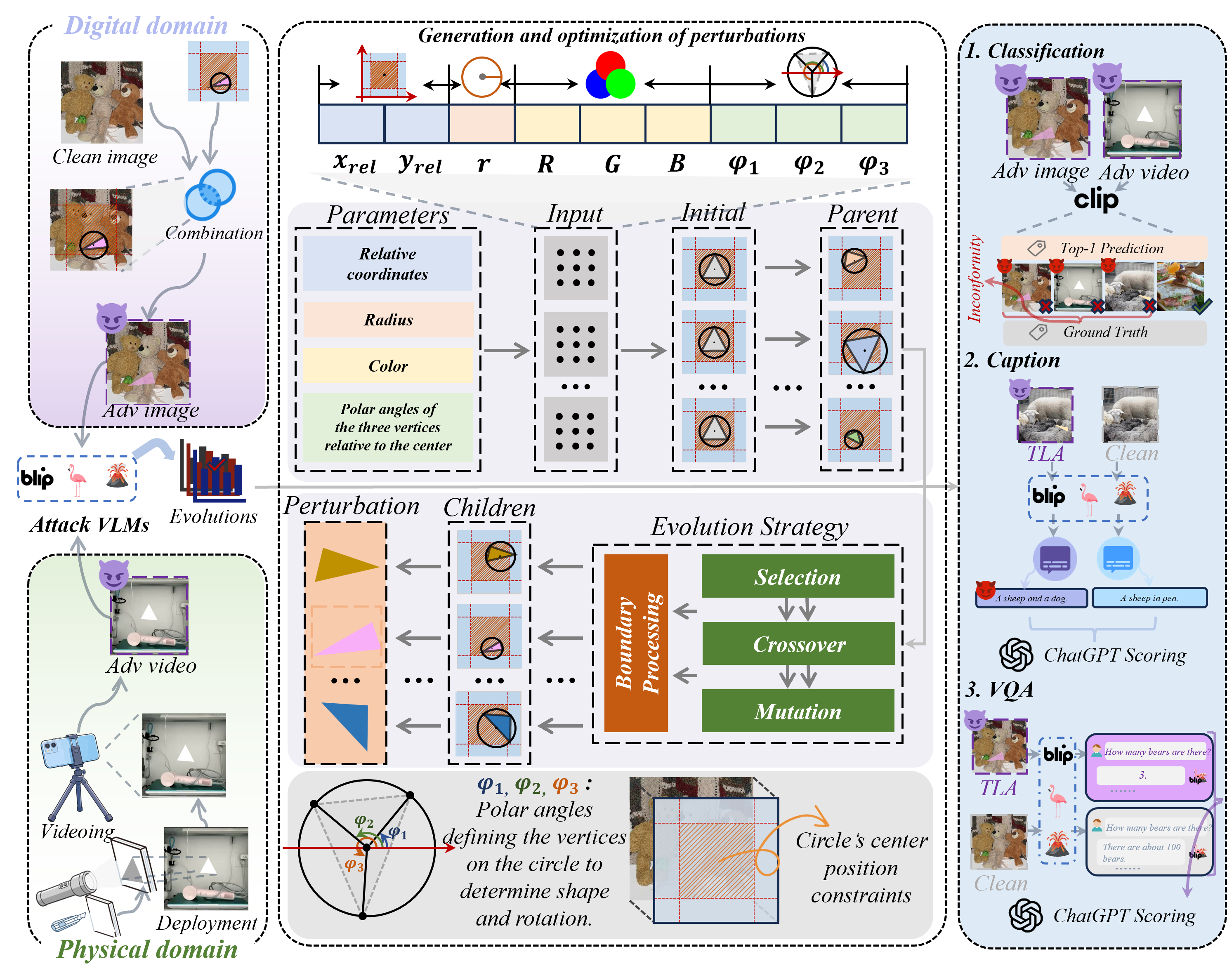}
  \caption{Overall framework of MSLA. Given a clean image, we parameterize the triangular light by its relative position, radius, color, and three polar angles, and optimize these variables with a genetic algorithm. The optimized triangular light can be deployed using simple optical devices such as a flashlight, colored transparent sheets, and shape templates.}
  \label{fig:MSLA-framework}
  \Description{An overview of the proposed MSLA framework.}
\end{figure*}

\section{Methodology}
In this section, we introduce the proposed MSLA framework, as shown in Figure~\ref{fig:MSLA-framework}. The MSLA framework uses a parametric triangular light pattern to simulate local lighting changes and generate adversarial samples.

\subsection{Problem Definition}

Given an input image $I$ and a candidate text label set $\mathcal{T}$,
where $\mathcal{T}=\{t_1,t_2,\ldots,t_k\}$ and the ground-truth label is $\bar{t} \in \mathcal{T}$, a vision-language model $F$ consists of an image encoder $E_I$ and a text encoder $E_T$. The model predicts the semantic alignment between image and text by computing cross-modal similarities:
\begin{equation}
z_k = s(E_I(I), E_T(t_k)), \quad k=1,2,\dots,K,
\end{equation}
where $s(\cdot,\cdot)$ denotes the similarity function in the joint embedding space. The prediction probability over all candidate labels is then obtained through a softmax operation:
\begin{equation}
p(t_k \mid I)=\frac{\exp(z_k)}{\sum_{j=1}^{K}\exp(z_j)}.
\end{equation}
Under normal conditions, a model is expected to correctly align the input image with its ground-truth text label:
\begin{equation}
\arg\max_{t_k \in \mathcal{T}} p(t_k \mid I)=\bar{t}.
\end{equation}

Our goal is to fool the vision-language model by applying a physically realizable localized triangular light perturbation to the image, thereby suppressing the confidence of the ground-truth label and disrupting the cross-modal alignment. Unlike conventional pixel-wise adversarial perturbations that modify the entire image, our attack is constrained to a localized light region that can be physically projected in the real world. Therefore, the adversarial example is formulated as: 
\begin{equation}
I^{adv}=\mathcal{A}(I,\Theta,\alpha,\mathcal{M}),
\end{equation}
where $\mathcal{A}(\cdot)$ denotes the light application function, $\Theta$ is the parameter set of the triangular light, $\alpha$ controls its transparency, and the binary mask $\mathcal{M}$ indicates whether a position is allowed to be modified, where one denotes allowed and zero denotes not allowed.

The objective of MSLA is to search for an optimal parameter set $\Theta^*$, where the resulting adversarial image minimizes the confidence assigned to the ground-truth label:
\begin{equation}
\Theta^*=\arg\min_{\Theta} p(\bar{t} \mid \mathcal{A}(I,\Theta,\alpha,\mathcal{M})).
\end{equation}

\subsection{Triangular Light Modeling}
In principle, the light pattern could be extended to more general polygonal shapes. However, we choose triangles because they are the simplest polygonal structure that still preserves sufficient shape diversity for adversarial modeling, while remaining natural in appearance and easy to realize in practice. Instead of directly optimizing three free vertices, which requires complicated geometric constraints to guarantee a valid triangle, we adopt a circle-based geometric modeling framework. Specifically, we first define a circle by its center and radius, and then generate three vertices on the circumference through angular parameters.

\noindent\textbf{Radius.} For an input image of fixed size $H \times W$, We select a radius $r$ from $[10, \gamma \min(H, W)]$, where $r$ is measured in pixels. A small radius may cause the three vertices to collapse onto the same pixel after coordinate rounding, degenerating the triangle into a single point. Therefore, the minimum radius is set to $10$ px. We further introduce a scaling factor $\gamma$ to constrain the maximum radius. As discussed in Section~\ref{sec:ablation}, we set $\gamma = 0.2$ to balance attack effectiveness and visual stealthiness, thereby preventing the light pattern from occupying an excessively large image region.

\noindent\textbf{Center.} To ensure that the generated triangle lies entirely within the valid image region, the circle center $o(x,y)$ is constrained within the feasible range determined by the current radius $r$. Specifically, the center coordinates are computed as:
\begin{equation}
\left\{
\begin{aligned}
x &= x_{\min} + x_{\mathrm{rel}} \cdot (x_{\max} - x_{\min}), \\
y &= y_{\min} + y_{\mathrm{rel}} \cdot (y_{\max} - y_{\min}),
\end{aligned}
\right.
\end{equation}

Here, $x_{\mathrm{rel}}$ and $y_{\mathrm{rel}}$ are relative coordinate parameters with values in $[0,1]$, which describe the offset of the circle center within the feasible region determined by $x_{\min}=r$, $x_{\max}=W-r$, $y_{\min}=r$, and $y_{\max}=H-r$. 

\noindent\textbf{Shape.} Unlike RFLA~\cite{ref19}, which can only generate a specific type of right triangle, this work aims to optimize a more general form of triangular light. To generate triangles of arbitrary shapes, we randomly choose three independent angles $\phi_1,\phi_2,\phi_3 \in [0, 360^\circ]$. The three distinct vertices $c_i(x_i,y_i)$ on the circumference are then determined by the following equations:

\begin{equation}
\left\{
\begin{aligned}
x_i &= x + r \times \sin\left(\phi_i \times \frac{\pi}{180}\right), \\
y_i &= y + r \times \cos\left(\phi_i \times \frac{\pi}{180}\right),
\end{aligned}
\right.
\quad i=1,2,3.
\end{equation}

\noindent\textbf{Color.} Besides geometry, the fill color of the triangular light is also incorporated into the attack modeling. In the digital setting, we randomly select a color tuple $(R,G,B)$ from the region of $[0,255]$.

\subsection{Adversarial Optimization}

Based on the above modeling, we develop an optimization strategy based on genetic algorithm to search for the optimal triangular light configuration.

\noindent\textbf{Optimization objective.}
The objective of MSLA is to find an adversarial light configuration that suppresses the confidence of the ground-truth label while increasing the uncertainty of model prediction. To this end, we define the fitness function as:
\begin{equation}
\mathcal{L}(\Theta)=\log\big(p(\bar{t} \mid I^{adv})+\epsilon\big)-\mathcal{H}(p(\cdot \mid I^{adv})),
\end{equation}
where $p(\bar{t} \mid I^{adv})$ denotes the probability assigned to the original ground-truth label, $\epsilon$ is a small constant (set to $e^{-10}$ in this paper) introduced to avoid numerical instability in the logarithm when $p(\bar{t} \mid I^{adv})$ approaches zero, and $\mathcal{H}(\cdot)$ is the Shannon entropy of the output distribution:
\begin{equation}
\mathcal{H}(p) = -\sum_{k=1}^{K} p(t_k \mid I^{adv}) \log p(t_k \mid I^{adv}).
\end{equation}

The first term suppresses the confidence of the correct label through logarithmic scaling. Compared with directly minimizing $p(\bar{t} \mid I^{adv})$, $\log\big(p(\bar{t} \mid I^{adv})+\epsilon\big)$ remains sensitive even when the ground-truth confidence is small. The second term increases the entropy of the prediction distribution, encouraging decision confusion and dispersing probability mass toward incorrect classes. Therefore, the optimal adversarial parameter set is obtained by:

\begin{equation}
\Theta^*=\arg\min_{\Theta}\mathcal{L}(\Theta).
\end{equation}

\noindent\textbf{Genetic algorithm.} Since the parameter vector $\Theta$ contains heterogeneous variables, including bounded continuous coordinates, color attributes, and angular parameters, the optimization landscape is highly non-convex and unsuitable for conventional gradient-based methods. To address this issue, we employ a genetic algorithm~\cite{ref20} to perform global search in the parameter space. Specifically, each individual in the population is encoded as a chromosome:
\begin{equation}
\Theta = (x_{\mathrm{rel}}, y_{\mathrm{rel}}, r, R, G, B, \phi_1, \phi_2, \phi_3).
\end{equation}
We first randomly initialize a population of candidate light configurations. In each generation, the fitness value of each individual is evaluated on the target model. Individuals with better attack performance, indicated by lower fitness values, are selected as parents through tournament selection. Then, crossover is applied to exchange partial parameter segments between parent chromosomes, enabling the search to explore new combinations of geometry and color. Mutation is further introduced on continuous variables such as center coordinates, radius, angles, and color channels, which increases population diversity and reduces the risk of getting trapped in local optima.
Through iterative selection, crossover, and mutation, the population gradually evolves toward more effective adversarial light configurations. The optimization terminates when either the model no longer predicts the ground-truth label as Top-1 or the maximum generation budget is reached.

\subsection{Physical Deployable Attack}

\begin{figure}[h]
  \centering
  \includegraphics[width=\linewidth]{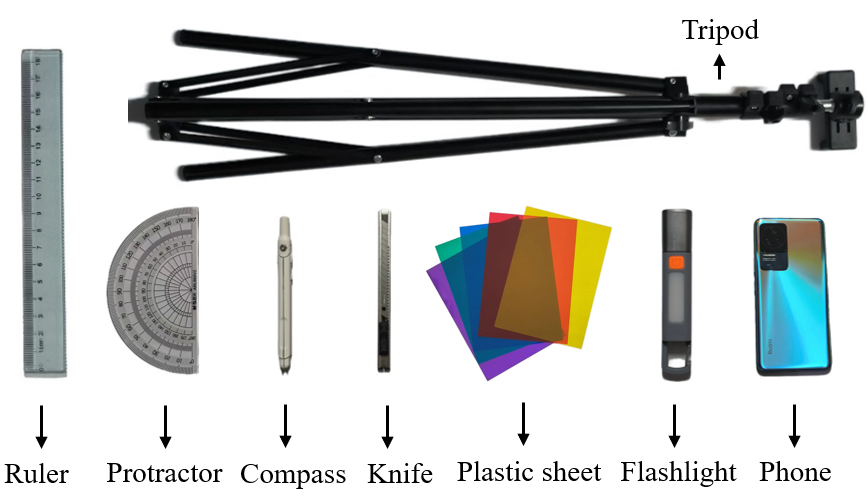}
  \caption{Experimental devices.}
  \label{fig:physical-equipment}
  \Description{The physical equipment used to deploy the MSLA.}
\end{figure}

In digital setting, we can generate colors by combining different RGB channel values. However, when conducting physical adversarial attacks, limited by equipment and physical materials, this search space is impractical. In the physical experiment, we collected the actual color of the selected transparent plastic sheet under the target camera. The sampled real physical color is taken as the initial search parameter of the genetic algorithm, rather than a randomly generated RGB sequence. At the same time, according to the optimized polar angles, we can locate the three angles through a protractor, quickly obtain the vertices of the triangle on the circle, and accurately cut out the shape template. The physical device is shown in Figure~\ref{fig:physical-equipment}.

\begin{table*}[t]
  \centering
  \caption{Performance Degradation ($\downarrow$) of VLMs on Zero-shot Classification Under Multimodal Semantic Lighting Attacks Examples Generated from the COCO Dataset. We compare the accuracy (\%) of existing illumination-based attack methods with our proposed method across four CLIP variants. Lower accuracy indicates stronger attack performance. The best result is shown in bold.}
  \label{tab:attack-comparison}
  \setlength{\tabcolsep}{5pt}
  \renewcommand{\arraystretch}{1.15}
  \resizebox{\textwidth}{!}{%
  \begin{tabular}{lcccc}
    \toprule
    \textbf{Method} 
    & \textbf{OpenCLIP ViT-B/16} 
    & \textbf{Meta-CLIP ViT-L/14} 
    & \textbf{EVA-CLIP ViT-G/14} 
    & \textbf{OpenAI CLIP ViT-L/14} \\
    \midrule
    Clean 
    & 97 
    & 98 
    & 98 
    & 93 \\
    
    Natural Light Attack~\cite{ref48}
    & 94 ({\color{red}$\downarrow 3$})
    & 97 ({\color{red}$\downarrow 1$})
    & 97 ({\color{red}$\downarrow 1$})
    & 93(-) \\
    
    Shadow Attack~\cite{ref47}
    & 84 ({\color{red}$\downarrow 13$})
    & 82 ({\color{red}$\downarrow 16$})
    & 95 ({\color{red}$\downarrow 3$})
    & 79 ({\color{red}$\downarrow 14$}) \\
    
    ITA~\cite{ref15}
    & 46 ({\color{red}$\downarrow 51$})
    & 64 ({\color{red}$\downarrow 34$})
    & 84 ({\color{red}$\downarrow 14$})
    & 51 ({\color{red}$\downarrow 42$}) \\
    
    \textbf{MSLA (ours)}
    & \textbf{29 ({\color{red}$\downarrow 68$})}
    & \textbf{23 ({\color{red}$\downarrow 75$})}
    & \textbf{36 ({\color{red}$\downarrow 62$})}
    & \textbf{11 ({\color{red}$\downarrow 82$})} \\
    \bottomrule
  \end{tabular}%
  }
\end{table*}

\section{Experiment}
To validate the effectiveness of the proposed MSLA, we conducted systematic digital and physical experiments. Moreover, to ensure the comparability and fairness of the digital evaluation, we strictly followed the experimental protocol established in ITA~\cite{ref15}.

\subsection{Experimental Setup}

\noindent\textbf{Datasets.}
We conduct our evaluation on the COCO dataset~\cite{ref45}. COCO provides diverse real-world scenes and object categories and has been widely adopted for evaluating the robustness of VLMs. We use the identical set of 300 images as ITA~\cite{ref15}, with each of the 30 categories represented by 10 images. We further adopt the 80 COCO categories as the semantic label list.

\noindent\textbf{Target models.}
We use the same experimental settings as ITA~\cite{ref15}. For the image classification task, we select four representative CLIP variants as attack targets, including OpenCLIP ViT-B/16~\cite{ref1,ref27}, Meta-CLIP ViT-L/14~\cite{ref28}, EVA-CLIP ViT-G/14~\cite{ref29}, and OpenAI CLIP ViT-L/14~\cite{ref30}. These models differ in parameter scale, pre-training data, and visual feature extraction capability, thereby enabling a comprehensive evaluation of the robustness of our method. For image captioning and visual question answering (VQA), we consider models with diverse architectures, including LLaVA-1.5~\cite{ref6}, LLaVA-1.6~\cite{ref31}, OpenFlamingo~\cite{ref32}, BLIP-2 ViT-L~\cite{ref21}, BLIP-2 FlanT5-XL~\cite{ref21}, and InstructBLIP~\cite{ref6}. We compare the proposed method against the following baselines: Shadow Attack~\cite{ref47}, Natural Light Attack~\cite{ref48}, and ITA~\cite{ref15}. As the current state-of-the-art illumination-based attack against vision-language models, ITA serves as the primary baseline in our study.

\noindent\textbf{Evaluation Metrics.}
For the image classification task, we report the attack success rate and top-1 accuracy, and further evaluate the relative performance drop compared to clean samples. For image captioning and VQA, we adopt the LLM-as-a-judge evaluation framework in accordance with recent studies~\cite{ref6,ref30,ref26}.

\noindent\textbf{Implementation details.}
We employ a genetic algorithm as the core optimization strategy to heuristically search the predefined parameter space over the coordinates, color, and the polar angles. In the optimization setup, the population size is set to 50 and the maximum number of generations is set to 200, while the crossover rate and mutation rate are fixed at 0.8 and 0.1, respectively. In particular, based on the quantitative findings of the ablation study in Section~\ref{sec:ablation}, we fix the transparency of the triangular light spot to $\alpha = 0.5$ and set the radius scaling factor to $0.2$ in order to maintain a balance between attack effectiveness and visual stealthiness. All attack experiments are conducted on an NVIDIA RTX 3090 (24GB) GPU.

\begin{table*}[h]
  \centering
  \caption{Performance Degradation ($\downarrow$) of VLMs on Image Captioning Under Multimodal Semantic Lighting Attacks Examples Generated from the COCO Dataset. We compare the consistency (\%) evaluated by GPT-4 of previous methods with our proposed approach. Lower values indicate better attack performance. The best result is shown in bold.}
  \label{tab:image-captioning}
  \setlength{\tabcolsep}{4pt}
  \renewcommand{\arraystretch}{1.1}
  \resizebox{\textwidth}{!}{%
  \begin{tabular}{llcccccc}
    \toprule
    \textbf{Image Encoder} & \textbf{Models} & \textbf{\#Params} & \textbf{Clean} & \textbf{Natural Light Attack~\cite{ref48}} & \textbf{Shadow Attack~\cite{ref47}} & \textbf{ITA~\cite{ref15}} & \textbf{MSLA (ours)} \\
    \midrule
    \multirow{4}{*}{OpenAI CLIP ViT-L/14}
      & LLaVA-1.5              & 7B   & 78.60 & 77.00({\color{red}$\downarrow 1.63$}) & 74.60({\color{red}$\downarrow 4.00$}) & 63.73({\color{red}$\downarrow 14.87$}) & \textbf{56.23({\color{red}$\downarrow 22.37$})} \\
      & LLaVA-1.6              & 7B   & 72.10 & 71.70({\color{red}$\downarrow 0.39$}) & 71.17({\color{red}$\downarrow 0.93$}) & 61.60({\color{red}$\downarrow 10.51$}) & \textbf{56.33({\color{red}$\downarrow 15.77$})} \\
      & OpenFlamingo           & 3B   & 70.20 & 69.53({\color{red}$\downarrow 0.67$}) & 67.80({\color{red}$\downarrow 2.40$}) & 53.93({\color{red}$\downarrow 16.27$}) & \textbf{41.73({\color{red}$\downarrow 28.47$})} \\
      & Blip-2 (FlanT5$_{\mathrm{XL}}$ ViT-L) & 3.4B & 75.10 & 70.77({\color{red}$\downarrow 4.33$}) & 68.57({\color{red}$\downarrow 6.53$}) & 60.93({\color{red}$\downarrow 14.17$}) & \textbf{53.88({\color{red}$\downarrow 21.22$})} \\
    \midrule
    \multirow{2}{*}{EVA-CLIP ViT-G/14}
      & Blip-2 (FlanT5$_{\mathrm{XL}}$) & 4.1B & 74.96 & 71.27({\color{red}$\downarrow 3.69$}) & 68.80({\color{red}$\downarrow 6.17$}) & 62.01({\color{red}$\downarrow 11.88$}) & \textbf{49.89({\color{red}$\downarrow 25.07$})} \\
      & InstructBLIP (FlanT5$_{\mathrm{XL}}$) & 4.1B & 76.50 & 72.07({\color{red}$\downarrow 4.43$}) & 69.77({\color{red}$\downarrow 6.73$}) & 63.20({\color{red}$\downarrow 13.30$}) & \textbf{59.43({\color{red}$\downarrow 17.07$})} \\
    \bottomrule
  \end{tabular}%
  }
\end{table*}

\begin{table*}[t]
  \centering
  \caption{Performance Degradation ($\downarrow$) of VLMs on VQA Task Under Multimodal Semantic Lighting Attacks Examples Generated from the COCO Dataset. We compare the correctness (\%) evaluated by GPT-4 of previous methods with our proposed approach. The performance is considered better with lower values. Number in bold indicates the best performance.}
  \label{tab:vqa}
  \setlength{\tabcolsep}{4pt}
  \renewcommand{\arraystretch}{1.1}
  \resizebox{\textwidth}{!}{%
  \begin{tabular}{llccccccc}
    \toprule
    \textbf{Image Encoder} & \textbf{Models} & \textbf{\#Params} & \textbf{Clean} & \textbf{Natural Light Attack~\cite{ref48}} & \textbf{Shadow Attack~\cite{ref47}} & \textbf{ITA~\cite{ref15}} & \textbf{MSLA (ours)} \\
    \midrule
    \multirow{4}{*}{OpenAI CLIP ViT-L/14}
      & LLaVA-1.5 & 7B   & 68.00 & 68.00(-) & 67.00({\color{red}$\downarrow 1.00$}) & 48.00({\color{red}$\downarrow 20.00$}) & \textbf{24.00({\color{red}$\downarrow 44.00$})} \\
      & LLaVA-1.6 & 7B   & 64.00 & 63.00({\color{red}$\downarrow 1.00$}) & 64.00(-) & 43.00({\color{red}$\downarrow 21.00$}) & \textbf{26.00({\color{red}$\downarrow 38.00$})} \\
      & OpenFlamingo & 3B   & 45.00 & 39.00({\color{red}$\downarrow 6.00$}) & 44.00({\color{red}$\downarrow 1.00$}) & 19.00({\color{red}$\downarrow 26.00$}) & \textbf{3.00({\color{red}$\downarrow 42.00$})} \\
      & Blip-2 (FlanT5$_{\mathrm{XL}}$ ViT-L) & 3.4B & 63.00 & 58.00({\color{red}$\downarrow 5.00$}) & 50.00({\color{red}$\downarrow 13.00$}) & 38.00({\color{red}$\downarrow 25.00$}) & \textbf{6.00({\color{red}$\downarrow 57.00$})} \\
    \midrule
    \multirow{2}{*}{EVA-CLIP ViT-G/14}
      & Blip-2 (FlanT5$_{\mathrm{XL}}$) & 4.1B & 54.00 & 53.00({\color{red}$\downarrow 1.00$}) & 54.00(-) & 33.00({\color{red}$\downarrow 21.00$}) & \textbf{10.00({\color{red}$\downarrow 44.00$})} \\
      & InstructBLIP (FlanT5$_{\mathrm{XL}}$) & 4.1B & 68.00 & 64.00({\color{red}$\downarrow 4.00$}) & 62.00({\color{red}$\downarrow 6.00$}) & 44.00({\color{red}$\downarrow 24.00$}) & \textbf{15.00({\color{red}$\downarrow 53.00$})} \\
    \bottomrule
  \end{tabular}%
  }
\end{table*}


\subsection{Digital Domain Performance}

\begin{figure}[h]
  \centering
  \includegraphics[width=\linewidth]{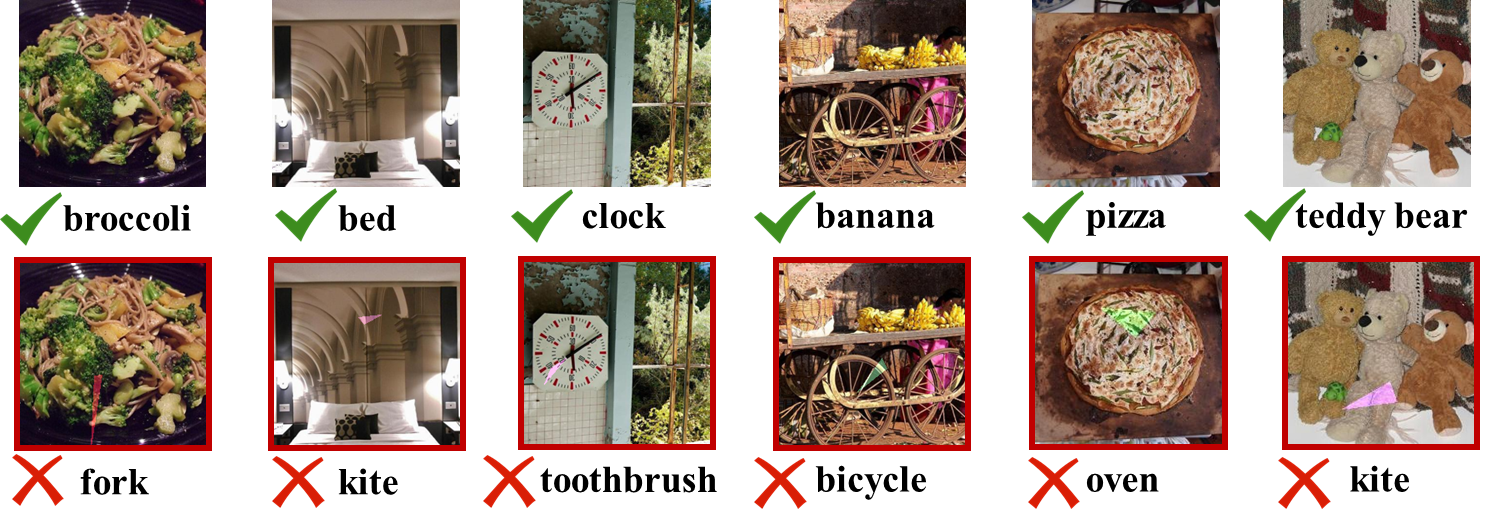}
  \caption{Visualization Results of Zero-shot Classification. The first row shows clean samples with their corresponding Top-1 labels, while the second row presents the MSLA samples with misclassified labels.}
  \label{fig:digital-examples}
  \Description{Representative adversarial examples generated by the proposed MSLA in the digital setting.}
\end{figure}

\noindent\textbf{Evaluation on Zero-Shot Classification.}
Table~\ref{tab:attack-comparison} compares our method with previous illumination-based attack methods on zero-shot classification tasks. Although the proposed attack is restricted to a local triangular region, it still induces substantial performance degradation across all target models. Our method consistently outperforms competing approaches, including ITA, on all CLIP variants. In particular, the top-1 accuracy on perturbed samples drops by nearly $60\%$ relative to clean samples for OpenCLIP ViT-B/16, EVA-CLIP ViT-G/14, and Meta-CLIP ViT-L/14, while the attack success rate on OpenAI CLIP ViT-L/14 reaches as high as $82\%$. The adversarial examples generated by our method are illustrated in Figure~\ref{fig:digital-examples}. It can be observed that VLMs can correctly recognize and classify the clean samples under normal conditions. However, after introducing the triangular light perturbation, the models’ predictions exhibit clear deviations and result in misclassification. 

\begin{figure*}[h]
  \centering
  \includegraphics[width=\textwidth]{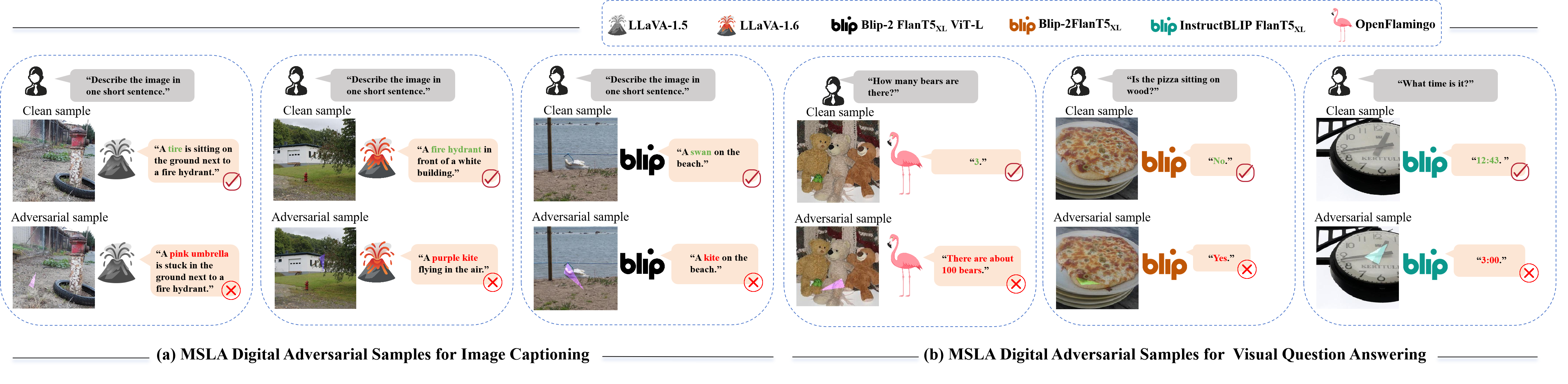}
  \caption{Visualization of some digital MSLA samples causing LVLMs to give incorrect answers. In each LVLM response, we highlight the {\color{green}correct} and {\color{red}error} parts using different colors.}
  \label{fig:vqa-captioning-examples}
  \Description{Representative adversarial examples showing incorrect outputs generated by vision-language models on image captioning and VQA tasks under the proposed MSLA.}
\end{figure*}



\noindent\textbf{Evaluation on Image Captioning Tasks.}
Beyond zero-shot classification, we further evaluate the effectiveness of the proposed MSLA on image captioning and VQA tasks. To ensure a fair and comprehensive comparison, we perform transfer attacks on large-scale Vision-Language Models (LVLMs) using samples generated by image encoders from different VLMs for zero-shot classification tasks. As shown in Table~\ref{tab:image-captioning}, the proposed MSLA consistently achieves the strongest attack effect on image captioning, yielding the lowest caption consistency scores across all evaluated models. Notably, smaller models suffer more severe degradation under the proposed attack. Smaller models such as OpenFlamingo and BLIP-2 exhibit substantial performance drops of approximately $28\%$ and $21\%$, respectively. In contrast, the larger LLaVA-1.6 model shows a relatively smaller degradation of about $15\%$, suggesting that models with larger parameter scales may possess stronger robustness against light perturbations. As illustrated in Figure~\ref{fig:vqa-captioning-examples} (a), while the models generate semantically accurate captions for clean images, they exhibit clear object hallucinations and semantic drift under attack, for example, misinterpreting a tire as a pink umbrella or a swan as a kite. These observations suggest that MSLA effectively disrupts visual grounding during caption generation.
\begin{figure}[h]
  \centering
  \includegraphics[width=\linewidth]{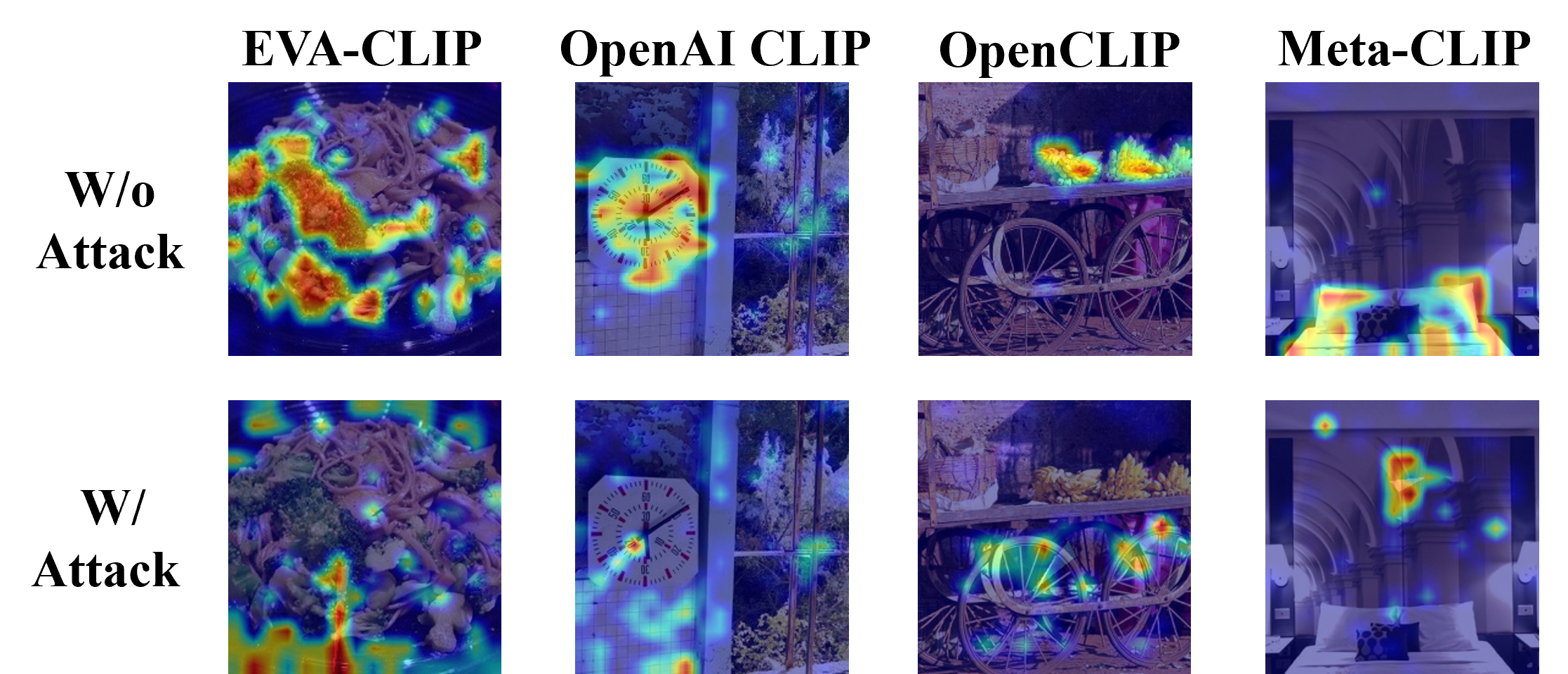}
  \caption{Attention analysis of CLIP models before and after our proposed method in the digital setting.}
  \label{fig:attention-analysis}
  \Description{Attention heatmaps of different CLIP variants on clean and adversarial samples, showing how the proposed MSLA shifts or diffuses model attention.}
\end{figure}

\begin{figure*}[t]
  \centering
  \includegraphics[width=\textwidth]{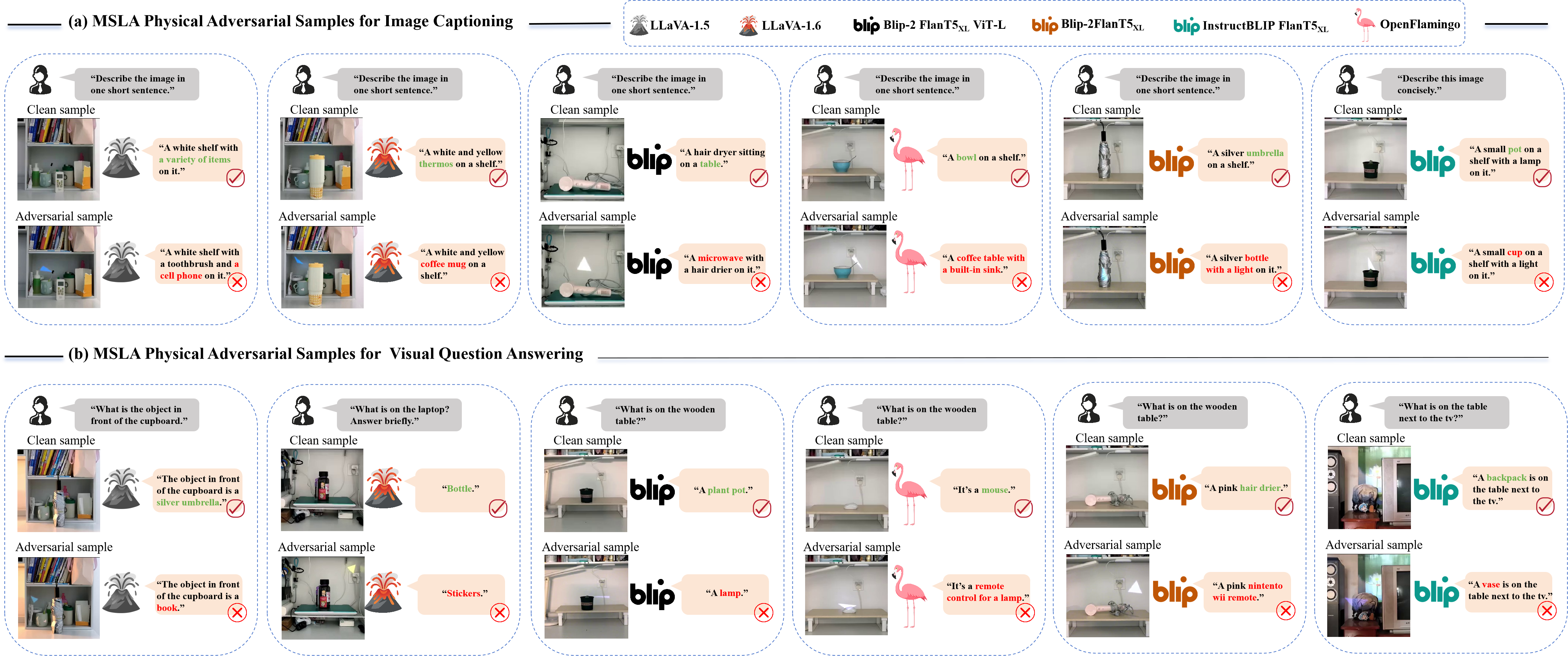}
  \caption{Visualization of physical MSLA samples causing LVLMs to give incorrect answers. In each LVLM response, we highlight the {\color{green}correct} and {\color{red}error} parts using different colors.}
  \label{fig:physical-vqa-captioning}
  \Description{Representative key frames from adversarial videos in the physical world, showing distorted captions and incorrect VQA answers generated by vision-language models under the proposed MSLA.}
\end{figure*}

\noindent\textbf{Evaluation on VQA Tasks.}
As reported in Table~\ref{tab:vqa}, MSLA is even more destructive on the VQA task, where it consistently outperforms all baselines by a large margin. Nearly all target models suffer substantial performance drops on VQA under attack, with correctness scores reduced by around $40\%$ overall regardless of model size. This indicates that reasoning-intensive tasks are especially vulnerable to localized light perturbations, as VQA depends more heavily on precise visual grounding and fine-grained integration of visual and textual information. The visual examples in Figure~\ref{fig:vqa-captioning-examples} (b) further illustrate that the attack causes failures in counting, spatial understanding, and visual reasoning. For instance, the model changes the correct counting answer from “3” to “There are about 100 bears,” flips the binary response to the question “Is the pizza sitting on wood?” from “No” to “Yes,” or misreads the clock time from “12:43” to “3:00.” These findings indicate that the perturbation propagates from corrupted visual perception to downstream cross-modal reasoning, leading to severe answer distortion.

\noindent\textbf{Attention Analysis.}
To further show the impact of the proposed MSLA on VLMs, we use Grad-CAM~\cite{ref24} to visualize the changes in attention heatmaps of four mainstream CLIP variants before and after the attack. For clean samples, the models' attention is accurately concentrated on the core semantic regions of the target objects. As shown in Figure~\ref{fig:attention-analysis}, OpenCLIP focuses on the banana, while MetaCLIP attends to the bed. However, after MSLA is applied, OpenCLIP shifts its attention from the banana to the wheel of a cart, leading to a misclassification as bicycle. Similarly, MetaCLIP redirects its attention from the bed to the wall background and incorrectly predicts kite. These results indicate that the triangular illuminated region successfully attracts and disrupts the models' visual attention.

\subsection{Physical Domain Performance}

By placing a colored transparent plastic sheet and a paper cutout of a specific shape in front of a flashlight, we can physically project a triangular light pattern onto a specific region, thereby enabling a physically deployable attack. The experimental samples are objects belonging to the COCO-80 categories that can be collected in the real world. All evaluation experiments are completed through dynamic video sequences of about 5 to 7 seconds recorded by mobile phones fixed on a tripod.

\begin{figure}[h]
  \centering
  \includegraphics[width=\linewidth]{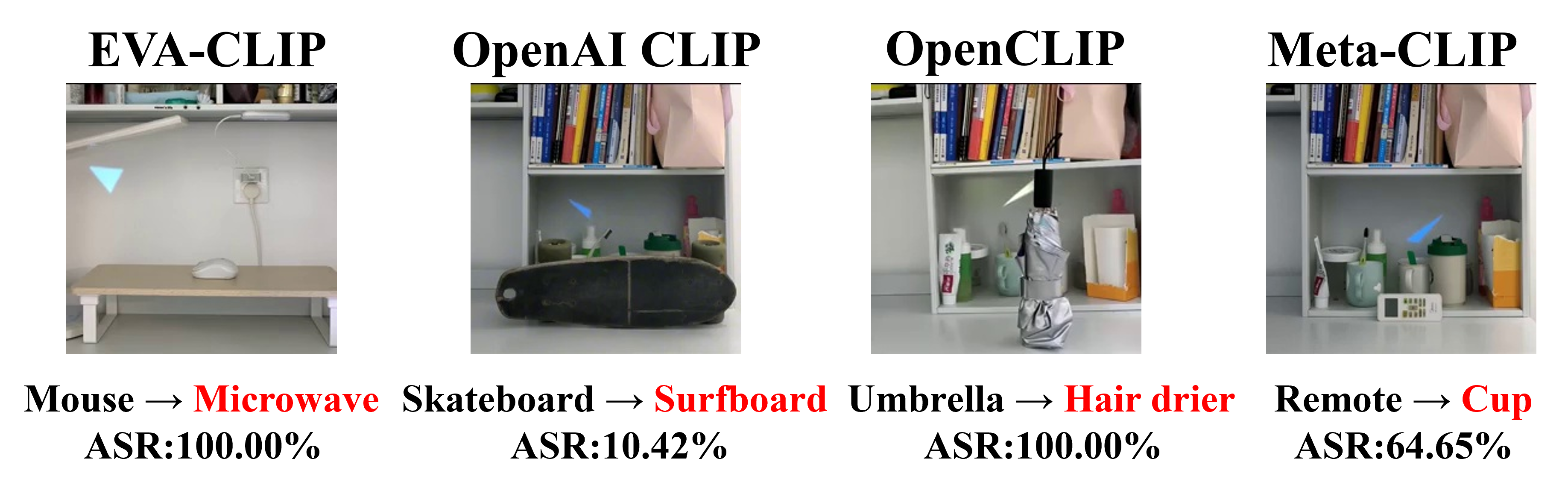}
  \caption{Physical MSLA samples for Zero-shot Classification with the corresponding frame-level ASR.}
  \label{fig:physical-classification}
  \Description{Representative frames extracted from adversarial videos in the physical-world zero-shot classification setting, showing successful attacks and corresponding prediction changes.}
\end{figure}

\noindent\textbf{Zero-Shot Classification in the Physical World.}
We perform optimization on the extracted clean video frames to obtain the triangular perturbation parameters that induce classification failure. Based on the optimized parameters, we prepare the corresponding cutout and use a transparent plastic sheet of the matched color to project the triangular light pattern onto the specific region, while recording an adversarial video. As shown in Figure~\ref{fig:physical-classification}, we present successful attack frames extracted from the adversarial videos together with the corresponding label changes. To quantitatively assess the stability of the attack in physical environments, we report the frame-level attack success rate, defined as the proportion of frames in the adversarial video that are misclassified by the model among all evaluated frames.

\begin{figure}[h]
  \centering
  \includegraphics[width=\linewidth]{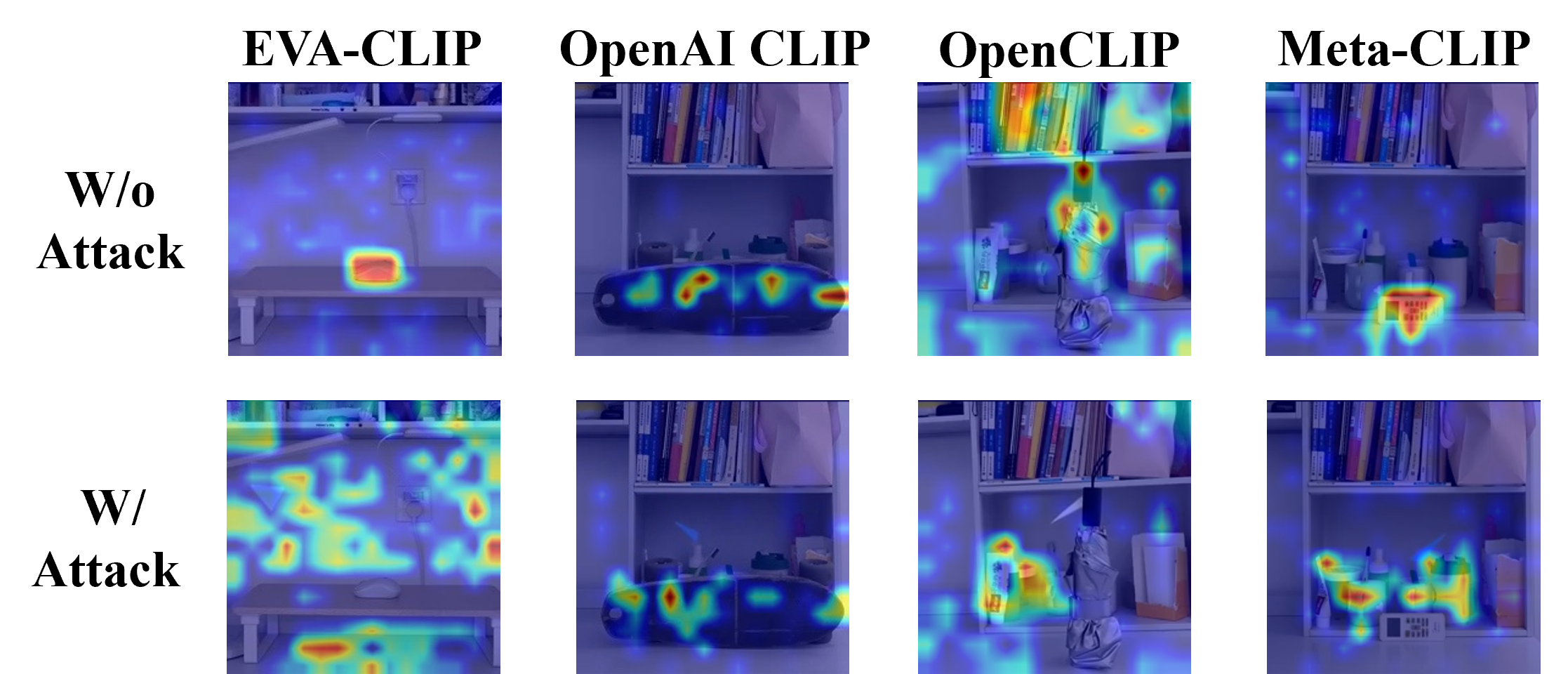}
  \caption{Attention analysis of CLIP models before and after our proposed method in the physical setting.}
  \label{fig:physical-attention}
  \Description{Attention heatmaps of aligned frames from clean and adversarial videos, showing changes in visual attention under the MSLA in the physical world.}
\end{figure}

\begin{figure*}[t]
  \centering
  \includegraphics[width=\textwidth]{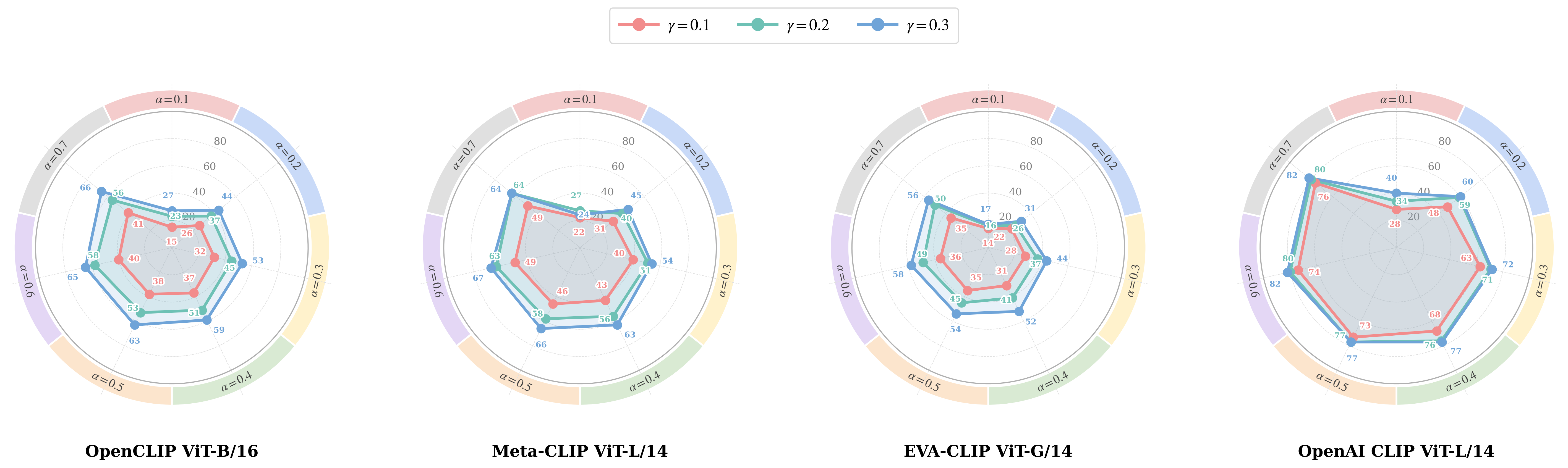}
  \caption{Attack performance under different transparency parameter $\alpha$ with varying radius scaling factors $\gamma$ on four CLIP variants.}
  \label{fig:ablation-alpha-gamma}
  \Description{Results of the ablation study showing attack performance under different transparency values and radius scaling factors.}
\end{figure*}

Meanwhile, we also visualize the attention heatmaps of aligned frames from the clean and adversarial videos, as shown in Figure~\ref{fig:physical-attention}. The results indicate that, for models with higher frame-level attack success rates, such as EVA-CLIP and OpenCLIP, the adversarial perturbation effectively causes visual attention to shift dramatically away from the target object. In contrast, for models with lower attack success rates, such as OpenAI CLIP, the attention regions remain largely consistent before and after the attack, without obvious dispersion or shift.

\noindent\textbf{Image Captioning Tasks in the Physical World.}
Similarly, based on the optimized parameters, we prepare the triangular cutout and use a flashlight together with a colored transparent sheet to project the light pattern while recording the adversarial video. For qualitative analysis, we extract key frames from the adversarial videos in which the LVLMs generate distorted captions. As shown in Figure~\ref{fig:physical-vqa-captioning} (a), the clean images can be described correctly by the target models, whereas under the physical triangular light perturbation, the generated captions deviate significantly from the actual scene. For example, in the clean sample, the scene is correctly described as ``a bowl on a shelf.'' However, under the physical triangular light perturbation, the same scene is miscaptioned as ``a coffee table with a built-in sink.'' The results indicate that our perturbations can disrupt the perception and generation processes of LVLMs in real physical environments.

\noindent\textbf{VQA Tasks in the Physical World.}
We further evaluate the proposed attack on VQA tasks in the physical world. As shown in Figure~\ref{fig:physical-vqa-captioning} (b), several mainstream LVLMs, including LLaVA, BLIP, and OpenFlamingo, are able to correctly answer object recognition and spatial reasoning questions based on clean images, such as ``What is the object in front of the cupboard?'' or ``What is on the wooden table?'' However, when exposed to triangular light with specific colors and positions, all of them generate incorrect answers that deviate from the actual physical scene. For example, the perturbed models may misidentify a backpack as a vase or a mouse as a remote control. These results further demonstrate that MSLA can effectively mislead VLMs in real physical environments through simple light-based perturbations.

\subsection{Ablation Study}
\label{sec:ablation}

\noindent\textbf{Attack Performance vs. Transparency and Radius Scaling Factor.}
The transparency parameter $\alpha$ controls the cover intensity of the color, while the radius scaling factor $\gamma$ determines its spatial extent. To isolate the influence of these two parameters, this ablation uses only the simplest confidence-based fitness function rather than the proposed multi-objective one.
As shown in Figure~\ref{fig:ablation-alpha-gamma}, the ASR on CLIP models increases with both $\alpha$ and $\gamma$. Specifically, ASR remains low when $\alpha < 0.3$, shows a substantial improvement at $\alpha = 0.5$, and gradually saturates as $\alpha$ further increases to $0.7$. Similarly, a small radius scaling factor ($\gamma = 0.1$) yields limited interference, whereas increasing $\gamma$ to $0.2$ significantly improves attack performance. Although a larger $\gamma$ of $0.3$ brings additional ASR gains, it also makes the perturbation resemble a broad global occlusion, reducing stealthiness. Therefore, considering both adversarial effectiveness and visual plausibility, we set the attack configuration to $\alpha = 0.5$ and $\gamma = 0.2$ in the main experiments.

\begin{table}[h]
  \centering
  \footnotesize
  \caption{Performance Degradation ($\downarrow$) of CLIP variants on Zero-shot Classification under different fitness functions. Lower accuracy (\%) indicates stronger attack performance. The best result in each row is shown in bold.}
  \label{tab:fitness_function_transposed}
  \setlength{\tabcolsep}{4pt}
  \renewcommand{\arraystretch}{1.2}
  \resizebox{\columnwidth}{!}{%
  \begin{tabular}{lccc}
    \toprule
    \textbf{CLIP variants} & \textbf{$\mathcal{L}_{\mathrm{prob}}(\Theta)$} & \textbf{$\mathcal{L}_{\mathrm{log}}(\Theta)$} & \textbf{$\mathcal{L}_{\mathrm{multi}}(\Theta)$ (ours)} \\
    \midrule
    OpenCLIP ViT-B/16   & 44({\color{red}$\downarrow 53$}) & 31({\color{red}$\downarrow 66$}) & \textbf{29({\color{red}$\downarrow 68$})} \\
    Meta-CLIP ViT-L/14  & 40({\color{red}$\downarrow 58$}) & 28({\color{red}$\downarrow 70$}) & \textbf{23({\color{red}$\downarrow 75$})} \\
    EVA-CLIP ViT-G/14   & 53({\color{red}$\downarrow 45$}) & 39({\color{red}$\downarrow 59$}) & \textbf{36({\color{red}$\downarrow 62$})} \\
    OpenAI CLIP ViT-L/14 & 16({\color{red}$\downarrow 77$}) & 13({\color{red}$\downarrow 80$}) & \textbf{11({\color{red}$\downarrow 82$})} \\
    \bottomrule
  \end{tabular}%
  }
\end{table}

\noindent\textbf{Attack Performance vs. Fitness Function.}
To validate the effectiveness of the proposed fitness function in the optimization process, we compare three different fitness function designs. As shown in Table~\ref{tab:fitness_function_transposed}, across all four CLIP variants our proposed multi-objective fitness function consistently achieves the strongest attack performance, inducing the largest performance drop on every model. In comparison, the attack based on the pure logit suppression objective performs slightly worse, while the variant using only the basic probability term $p(\bar{t} \mid I^{adv})$ as the fitness function yields the lowest results. These findings demonstrate that our fitness function can more effectively steer the genetic algorithm toward optimal light perturbation parameters, thereby maximizing the adversarial effect on VLMs.

\section{Conclusion}
In this work, we propose the MSLA framework to reveal and evaluate the security vulnerabilities of VLMs. We parameterize the geometric shape, position, color, and transparency of the triangular light, and optimize them with a genetic algorithm. Extensive experiments demonstrate that MSLA significantly degrade the performance of a wide range of mainstream VLMs across zero-shot image classification, image captioning, and VQA tasks. It can also be physically deployed with simple devices such as a flashlight and colored transparent plastic sheets. This study presents a new physical-world adversarial attack paradigm for VLMs while exposing their vulnerability to realistic illumination variations, emphasizing the need for stronger robustness against lighting disturbances.

\bibliographystyle{ACM-Reference-Format}
\bibliography{sample-base}

@String{Computing = "Computing" }

@String{Computer = "{IEEE} Computer" }

@String{Springer = "Springer-Verlag" }

@inproceedings{ref1,
  title={Learning transferable visual models from natural language supervision},
  author={Radford, Alec and Kim, Jong Wook and Hallacy, Chris and Ramesh, Aditya and Goh, Gabriel and Agarwal, Sandhini and Sastry, Girish and Askell, Amanda and Mishkin, Pamela and Clark, Jack and others},
  booktitle={International conference on machine learning},
  pages={8748--8763},
  year={2021},
  organization={PmLR}
}

@inproceedings{ref2,
  title={Reproducible scaling laws for contrastive language-image learning},
  author={Cherti, Mehdi and Beaumont, Romain and Wightman, Ross and Wortsman, Mitchell and Ilharco, Gabriel and Gordon, Cade and Schuhmann, Christoph and Schmidt, Ludwig and Jitsev, Jenia},
  booktitle={Proceedings of the IEEE/CVF conference on computer vision and pattern recognition},
  pages={2818--2829},
  year={2023}
}

@inproceedings{ref3,
  title={Unified vision-language pre-training for image captioning and vqa},
  author={Zhou, Luowei and Palangi, Hamid and Zhang, Lei and Hu, Houdong and Corso, Jason and Gao, Jianfeng},
  booktitle={Proceedings of the AAAI conference on artificial intelligence},
  volume={34},
  number={07},
  pages={13041--13049},
  year={2020}
}

@article{ref4,
  title={Flamingo: a visual language model for few-shot learning},
  author={Alayrac, Jean-Baptiste and Donahue, Jeff and Luc, Pauline and Miech, Antoine and Barr, Iain and Hasson, Yana and Lenc, Karel and Mensch, Arthur and Millican, Katherine and Reynolds, Malcolm and others},
  journal={Advances in neural information processing systems},
  volume={35},
  pages={23716--23736},
  year={2022}
}

@inproceedings{ref5,
  title={Internvl: Scaling up vision foundation models and aligning for generic visual-linguistic tasks},
  author={Chen, Zhe and Wu, Jiannan and Wang, Wenhai and Su, Weijie and Chen, Guo and Xing, Sen and Zhong, Muyan and Zhang, Qinglong and Zhu, Xizhou and Lu, Lewei and others},
  booktitle={Proceedings of the IEEE/CVF conference on computer vision and pattern recognition},
  pages={24185--24198},
  year={2024}
}

@article{ref6,
  title={Visual instruction tuning},
  author={Liu, Haotian and Li, Chunyuan and Wu, Qingyang and Lee, Yong Jae},
  journal={Advances in neural information processing systems},
  volume={36},
  pages={34892--34916},
  year={2023}
}

@misc{ref7,
  title={Llavanext: Improved reasoning, ocr, and world knowledge},
  author={Liu, Haotian and Li, Chunyuan and Li, Yuheng and Li, Bo and Zhang, Yuanhan and Shen, Sheng and Lee, Yong Jae},
  year={2024}
}

@article{ref8,
  title={Double domain guided real-time low-light image enhancement for ultra-high-definition transportation surveillance},
  author={Qu, Jingxiang and Liu, Ryan Wen and Gao, Yuan and Guo, Yu and Zhu, Fenghua and Wang, Fei-Yue},
  journal={IEEE Transactions on Intelligent Transportation Systems},
  volume={25},
  number={8},
  pages={9550--9562},
  year={2024},
  publisher={IEEE}
}

@article{ref9,
  title={Visual adversarial attack on vision-language models for autonomous driving},
  author={Zhang, Tianyuan and Wang, Lu and Zhang, Xinwei and Zhang, Yitong and Jia, Boyi and Liang, Siyuan and Hu, Shengshan and Fu, Qiang and Liu, Aishan and Liu, Xianglong},
  journal={arXiv preprint arXiv:2411.18275},
  year={2024}
}

@inproceedings{ref11,
  title={On the robustness of large multimodal models against image adversarial attacks},
  author={Cui, Xuanming and Aparcedo, Alejandro and Jang, Young Kyun and Lim, Ser-Nam},
  booktitle={Proceedings of the IEEE/CVF Conference on Computer Vision and Pattern Recognition},
  pages={24625--24634},
  year={2024}
}

@inproceedings{ref12,
  title={Towards viewpoint-invariant visual recognition via adversarial training},
  author={Ruan, Shouwei and Dong, Yinpeng and Su, Hang and Peng, Jianteng and Chen, Ning and Wei, Xingxing},
  booktitle={Proceedings of the IEEE/CVF International Conference on Computer Vision},
  pages={4709--4719},
  year={2023}
}

@article{ref13,
  title={State-of-the-art optical-based physical adversarial attacks for deep learning computer vision systems},
  author={Fang, Junbin and Jiang, You and Jiang, Canjian and Jiang, Zoe L and Liu, Chuanyi and Yiu, Siu-Ming},
  journal={Expert Systems with Applications},
  volume={250},
  pages={123761},
  year={2024},
  publisher={Elsevier}
}

@article{ref14,
  title={Unibench: Visual reasoning requires rethinking vision-language beyond scaling},
  author={Al-Tahan, Haider and Garrido, Quentin and Balestriero, Randall and Bouchacourt, Diane and Hazirbas, Caner and Ibrahim, Mark},
  journal={Advances in Neural Information Processing Systems},
  volume={37},
  pages={82411--82437},
  year={2024}
}

@inproceedings{ref15,
  title={When Lighting Deceives: Exposing Vision-Language Models' Illumination Vulnerability Through Illumination Transformation Attack},
  author={Liu, Hanqing and Ruan, Shouwei and Huang, Yao and Zhao, Shiji and Wei, Xingxing},
  booktitle={Proceedings of the IEEE/CVF International Conference on Computer Vision},
  pages={10485--10495},
  year={2025}
}

@article{ref16,
  title={Light as Deception: GPT-driven Natural Relighting Against Vision-Language Pre-training Models},
  author={Yang, Ying and Zhang, Jie and Lv, Xiao and Lin, Di and Xiang, Tao and Guo, Qing},
  journal={arXiv preprint arXiv:2505.24227},
  year={2025}
}

@inproceedings{ref17,
  title={Embodied laser attack: Leveraging scene priors to achieve agent-based robust non-contact attacks},
  author={Sun, Yitong and Huang, Yao and Wei, Xingxing},
  booktitle={Proceedings of the 32nd ACM International Conference on Multimedia},
  pages={5902--5910},
  year={2024}
}

@inproceedings{ref18,
  title={Optical adversarial attack},
  author={Gnanasambandam, Abhiram and Sherman, Alex M and Chan, Stanley H},
  booktitle={Proceedings of the IEEE/CVF international conference on computer vision},
  pages={92--101},
  year={2021}
}

@inproceedings{ref19,
  title={Rfla: A stealthy reflected light adversarial attack in the physical world},
  author={Wang, Donghua and Yao, Wen and Jiang, Tingsong and Li, Chao and Chen, Xiaoqian},
  booktitle={Proceedings of the IEEE/CVF international conference on computer vision},
  pages={4455--4465},
  year={2023}
}

@article{ref20,
  title={Genetic algorithms},
  author={Holland, John H},
  journal={Scientific american},
  volume={267},
  number={1},
  pages={66--73},
  year={1992},
  publisher={JSTOR}
}

@inproceedings{ref21,
  title={Blip-2: Bootstrapping language-image pre-training with frozen image encoders and large language models},
  author={Li, Junnan and Li, Dongxu and Savarese, Silvio and Hoi, Steven},
  booktitle={International conference on machine learning},
  pages={19730--19742},
  year={2023},
  organization={PMLR}
}

@article{ref23,
  title={Laion-5b: An open large-scale dataset for training next generation image-text models},
  author={Schuhmann, Christoph and Beaumont, Romain and Vencu, Richard and Gordon, Cade and Wightman, Ross and Cherti, Mehdi and Coombes, Theo and Katta, Aarush and Mullis, Clayton and Wortsman, Mitchell and others},
  journal={Advances in neural information processing systems},
  volume={35},
  pages={25278--25294},
  year={2022}
}

@inproceedings{ref24,
  title={Grad-cam: Visual explanations from deep networks via gradient-based localization},
  author={Selvaraju, Ramprasaath R and Cogswell, Michael and Das, Abhishek and Vedantam, Ramakrishna and Parikh, Devi and Batra, Dhruv},
  booktitle={Proceedings of the IEEE international conference on computer vision},
  pages={618--626},
  year={2017}
}

@article{ref25,
  title={Real-time object detection, tracking, and monitoring framework for security surveillance systems},
  author={Abba, Sani and Bizi, Ali Mohammed and Lee, Jeong-A and Bakouri, Souley and Crespo, Maria Liz},
  journal={Heliyon},
  volume={10},
  number={15},
  year={2024},
  publisher={Elsevier}
}

@inproceedings{ref26,
 author = {Zhang, Yichi and Huang, Yao and Sun, Yitong and Liu, Chang and Zhao, Zhe and Fang, Zhengwei and Wang, Yifan and Chen, Huanran and Yang, Xiao and Wei, Xingxing and Su, Hang and Dong, Yinpeng and Zhu, Jun},
 booktitle = {Advances in Neural Information Processing Systems},
 doi = {10.52202/079017-1561},
 editor = {A. Globerson and L. Mackey and D. Belgrave and A. Fan and U. Paquet and J. Tomczak and C. Zhang},
 pages = {49279--49383},
 publisher = {Curran Associates, Inc.},
 title = {MultiTrust: A Comprehensive Benchmark Towards Trustworthy Multimodal Large Language Models},
 url = {https://proceedings.neurips.cc/paper_files/paper/2024/file/586640cda3db2dc77349013dcefee456-Paper-Datasets_and_Benchmarks_Track.pdf},
 volume = {37},
 year = {2024}
}

@article{ref27,
  title={OpenCLIP: An open source implementation of CLIP},
  author={Ilharco, Gabriel and Wortsman, Mitchell and Carlini, Nicholas and Anil, Rohan and Minderer, Matthias and Zhai, Xiaohua and Dosovitskiy, Alexey and Beyer, Lucas and Talwar, Kunal and Steiner, Andreas and others},
  journal={GitHub repository},
  year={2021}
}

@article{ref28,
  title={Demystifying clip data},
  author={Xu, Hu and Xie, Saining and Tan, Xiaoqing Ellen and Huang, Po-Yao and Howes, Russell and Sharma, Vasu and Li, Shang-Wen and Ghosh, Gargi and Zettlemoyer, Luke and Feichtenhofer, Christoph},
  journal={arXiv preprint arXiv:2309.16671},
  year={2023}
}

@article{ref29,
  title={Eva-clip: Improved training techniques for clip at scale},
  author={Sun, Quan and Fang, Yuxin and Wu, Ledell and Wang, Xinlong and Cao, Yue},
  journal={arXiv preprint arXiv:2303.15389},
  year={2023}
}

@article{ref30,
  title={Judging llm-as-a-judge with mt-bench and chatbot arena},
  author={Zheng, Lianmin and Chiang, Wei-Lin and Sheng, Ying and Zhuang, Siyuan and Wu, Zhanghao and Zhuang, Yonghao and Lin, Zi and Li, Zhuohan and Li, Dacheng and Xing, Eric and others},
  journal={Advances in neural information processing systems},
  volume={36},
  pages={46595--46623},
  year={2023}
}

@inproceedings{ref31,
  title={Improved baselines with visual instruction tuning},
  author={Liu, Haotian and Li, Chunyuan and Li, Yuheng and Lee, Yong Jae},
  booktitle={Proceedings of the IEEE/CVF conference on computer vision and pattern recognition},
  pages={26296--26306},
  year={2024}
}

@article{ref32,
  title={Openflamingo: An open-source framework for training large autoregressive vision-language models},
  author={Awadalla, Anas and Gao, Irena and Gardner, Josh and Hessel, Jack and Hanafy, Yusuf and Zhu, Wanrong and Marathe, Kalyani and Bitton, Yonatan and Gadre, Samir and Sagawa, Shiori and others},
  journal={arXiv preprint arXiv:2308.01390},
  year={2023}
}

@article{ref34,
  title={Drivevlm: The convergence of autonomous driving and large vision-language models},
  author={Tian, Xiaoyu and Gu, Junru and Li, Bailin and Liu, Yicheng and Wang, Yang and Zhao, Zhiyong and Zhan, Kun and Jia, Peng and Lang, Xianpeng and Zhao, Hang},
  journal={arXiv preprint arXiv:2402.12289},
  year={2024}
}

@article{ref35,
  title={Visualbert: A simple and performant baseline for vision and language},
  author={Li, Liunian Harold and Yatskar, Mark and Yin, Da and Hsieh, Cho-Jui and Chang, Kai-Wei},
  journal={arXiv preprint arXiv:1908.03557},
  year={2019}
}

@article{ref36,
  title={Vilbert: Pretraining task-agnostic visiolinguistic representations for vision-and-language tasks},
  author={Lu, Jiasen and Batra, Dhruv and Parikh, Devi and Lee, Stefan},
  journal={Advances in neural information processing systems},
  volume={32},
  year={2019}
}

@article{ref37,
  title={On evaluating adversarial robustness of large vision-language models},
  author={Zhao, Yunqing and Pang, Tianyu and Du, Chao and Yang, Xiao and Li, Chongxuan and Cheung, Ngai-Man Man and Lin, Min},
  journal={Advances in Neural Information Processing Systems},
  volume={36},
  pages={54111--54138},
  year={2023}
}

@article{ref38,
  title={Vlattack: Multimodal adversarial attacks on vision-language tasks via pre-trained models},
  author={Yin, Ziyi and Ye, Muchao and Zhang, Tianrong and Du, Tianyu and Zhu, Jinguo and Liu, Han and Chen, Jinghui and Wang, Ting and Ma, Fenglong},
  journal={Advances in Neural Information Processing Systems},
  volume={36},
  pages={52936--52956},
  year={2023}
}

@inproceedings{ref39,
  title={Omniview-tuning: Boosting viewpoint invariance of vision-language pre-training models},
  author={Ruan, Shouwei and Dong, Yinpeng and Liu, Hanqing and Huang, Yao and Su, Hang and Wei, Xingxing},
  booktitle={European Conference on Computer Vision},
  pages={309--327},
  year={2024},
  organization={Springer}
}

@InProceedings{ref40,
    author    = {Tong, Shengbang and Liu, Zhuang and Zhai, Yuexiang and Ma, Yi and LeCun, Yann and Xie, Saining},
    title     = {Eyes Wide Shut? Exploring the Visual Shortcomings of Multimodal LLMs},
    booktitle = {Proceedings of the IEEE/CVF Conference on Computer Vision and Pattern Recognition (CVPR)},
    month     = {June},
    year      = {2024},
    pages     = {9568-9578}
}

@InProceedings{ref41,
    author    = {Duan, Ranjie and Mao, Xiaofeng and Qin, A. K. and Chen, Yuefeng and Ye, Shaokai and He, Yuan and Yang, Yun},
    title     = {Adversarial Laser Beam: Effective Physical-World Attack to DNNs in a Blink},
    booktitle = {Proceedings of the IEEE/CVF Conference on Computer Vision and Pattern Recognition (CVPR)},
    month     = {June},
    year      = {2021},
    pages     = {16062-16071}
}

@article{ref42,
  title={Understanding zero-shot adversarial robustness for large-scale models},
  author={Mao, Chengzhi and Geng, Scott and Yang, Junfeng and Wang, Xin and Vondrick, Carl},
  journal={arXiv preprint arXiv:2212.07016},
  year={2022}
}

@article{ref45,
  title={Microsoft coco captions: Data collection and evaluation server},
  author={Chen, Xinlei and Fang, Hao and Lin, Tsung-Yi and Vedantam, Ramakrishna and Gupta, Saurabh and Doll{\'a}r, Piotr and Zitnick, C Lawrence},
  journal={arXiv preprint arXiv:1504.00325},
  year={2015}
}

@inproceedings{ref46,
  title={Spaa: Stealthy projector-based adversarial attacks on deep image classifiers},
  author={Huang, Bingyao and Ling, Haibin},
  booktitle={2022 IEEE Conference on Virtual Reality and 3D User Interfaces (VR)},
  pages={534--542},
  year={2022},
  organization={IEEE}
}

@inproceedings{ref47,
  title={Shadows can be dangerous: Stealthy and effective physical-world adversarial attack by natural phenomenon},
  author={Zhong, Yiqi and Liu, Xianming and Zhai, Deming and Jiang, Junjun and Ji, Xiangyang},
  booktitle={Proceedings of the IEEE/CVF conference on computer vision and pattern recognition},
  pages={15345--15354},
  year={2022}
}

@inproceedings{ref48,
  title={Natural light can also be dangerous: Traffic sign misinterpretation under adversarial natural light attacks},
  author={Hsiao, Teng-Fang and Huang, Bo-Lun and Ni, Zi-Xiang and Lin, Yan-Ting and Shuai, Hong-Han and Li, Yung-Hui and Cheng, Wen-Huang},
  booktitle={Proceedings of the IEEE/CVF Winter Conference on Applications of Computer Vision},
  pages={3915--3924},
  year={2024}
}

@inproceedings{ref49,
  title={Towards adversarial attack on vision-language pre-training models},
  author={Zhang, Jiaming and Yi, Qi and Sang, Jitao},
  booktitle={Proceedings of the 30th ACM International Conference on Multimedia},
  pages={5005--5013},
  year={2022}
}

@inproceedings{ref50,
  title={Blip: Bootstrapping language-image pre-training for unified vision-language understanding and generation},
  author={Li, Junnan and Li, Dongxu and Xiong, Caiming and Hoi, Steven},
  booktitle={International conference on machine learning},
  pages={12888--12900},
  year={2022},
  organization={PMLR}
}

@article{ref51,
  title={Spatial-aware Vision Language Model for Autonomous Driving},
  author={Wei, Weijie and Luo, Zhipeng and Feng, Ling and Liong, Venice Erin},
  journal={arXiv preprint arXiv:2512.24331},
  year={2025}
}

@article{ref52,
  title={A survey of attacks on large vision--language models: Resources, advances, and future trends},
  author={Liu, Daizong and Yang, Mingyu and Qu, Xiaoye and Zhou, Pan and Cheng, Yu and Hu, Wei},
  journal={IEEE Transactions on Neural Networks and Learning Systems},
  year={2025},
  publisher={IEEE}
}

@inproceedings{ref53,
  title={On the Adversarial Robustness of Visual-Language Chat Models},
  author={Qin, Tianrui and Wang, Xuan and Zhao, Juanjuan and Ye, Kejiang and Xu, Cheng-zhong and Gao, Xitong},
  booktitle={Proceedings of the 2025 International Conference on Multimedia Retrieval},
  pages={1118--1127},
  year={2025}
}

@ArtifactSoftware{R,
    title = {R: A Language and Environment for Statistical Computing},
    author = {{R Core Team}},
    organization = {R Foundation for Statistical Computing},
    address = {Vienna, Austria},
    year = {2019},
    url = {https://www.R-project.org/},
}










\end{document}